\documentclass[runningheads]{llncs}
\usepackage{amsmath}
\usepackage{graphicx}

\begin{document}
\title{Human or Machine: Automating Human Likeliness Evaluation of NLG Texts\thanks{Supported by the project no. 19-26934X (NEUREM3) of the Czech Science Foundation and ELITR  (H2020-ICT-2018-2-825460) of the EU.}}
\titlerunning{Automating Human Likeliness Evaluation}
\author{Erion \c{C}ano\orcidID{0000-0002-5496-3860} \and
Ond\v{r}ej Bojar\orcidID{0000-0002-0606-0050}}
\authorrunning{E. \c{C}ano and O. Bojar}
\institute{Charles University in Prague, Czech Republic\\
\email{\{cano, bojar\}@ufal.mff.cuni.cz}}

\maketitle              %
\begin{abstract}
Automatic evaluation of various text quality criteria produced by data-driven intelligent methods is very common and useful because it is cheap, fast, and usually yields repeatable results. In this paper, we present an attempt to automate the human likeliness evaluation of the output text samples coming from natural language generation methods used to solve several tasks. We propose to use a human likeliness score that shows the percentage of the output samples from a method that look as if they were written by a human. Instead of having human participants label or rate those samples, we completely automate the process by using a discrimination procedure based on large pretrained language models and their probability distributions. As follow up, we plan to perform an empirical analysis of human-written and machine-generated texts to find the optimal setup of this evaluation approach. A validation procedure involving human participants will also check how the automatic evaluation correlates with human judgments.   
\keywords{Natural Language Generation \and Automatic Evaluation \and Human Likeliness \and Text Naturalness \and Evaluation Metrics.}
\end{abstract}
\section{Introduction}

NLG (Natural Language Generation) is a set of techniques and practices for automatically transforming structured data into natural language. Among the typical applications of NLG, we can mention weather predictions, financial reporting, customer support, etc. Same as other related disciplines such as MT (Machine Translation) or TS (Text Summarization), it has surged in the last decade, greatly pushed by the significant advances in text applications of deep neural networks \cite{8416973,Balaji2018} as well as the creation of large datasets \cite{K16-1028,cano-bojar-2020-two,Napoles:2012:AG:2391200.2391218}. In parallel with the data processing, researchers are also finding ways to automate the evaluation of the intelligent text-related systems they propose. 
An evaluation practice that is gaining popularity compares method output texts against human-written references of a standard corpus using automatic metrics. Some of the most popular metrics are BLEU of Papineni et al. \cite{P02-1040}, ROUGE of Lin \cite{Lin:2004}, and METEOR of Banerjee and Lavie \cite{banerjee-lavie-2005-meteor}. This automatic evaluation practice is tempting for NLP (and NLG) researchers since it is fast and cheap to perform, does not require domain expertise, and usually produces repeatable results. 
When comparing two methods \textbf{A} and \textbf{B}, besides accuracy, we usually want to also observe other quality aspects (e.g., coherence, readability, naturalness, fluency, adequacy or grammaticality) of the text they produce. BLEU and the automatic evaluation process have yet been criticized by several authors \cite{sulem-etal-2018-bleu,bojar-etal-2010-tackling}. As pointed out by Reiter \cite{reiter-2018-structured}, BLEU is not appropriate for evaluating all quality criteria of the NLG systems. Furthermore, the results of the automatic evaluation process do not always correlate well with those of human campaigns \cite{reiter-belz-2009-investigation}. Even adding the data efficiency scores in the process would not tell us anything about the text quality aspects of \textbf{A} and \textbf{B} outputs \cite{cano-bojar-2019-efficiency}.  
As pointed out by Novikova et al. \cite{novikova-etal-2017-need}, there is a need for new evaluation metrics to objectively assess specific text quality characteristics such as human likeliness (or naturalness), coherence, etc.  
In this paper, we present the proof of concepts for a procedure that can be used to automatically evaluate human likeliness of NLG outputs. We propose to use a discriminator model that can label each test set output as being human-written (natural) or machine-generated (synthetic). This model will use existing pretrained language models such as BERT or GPT-2 and assume that synthetic texts do mostly contain high-rank words sampled from probability distributions, in contrast to natural texts that contain more low-rank words. This way, it will compute the fraction probability (the ratio between the probability of the actual word in a position and the highest rank word for that position) of each word and the average value for the entire text sample. The latter will be discretized to get the sample class (\emph{h} or \emph{m}).  
We start with a survey of evaluation procedures in the last five INLG conference proceedings and notice an increasing trend of automatic evaluations against human-based ones. Later on, we discuss the possibility of automating human likeliness evaluation and propose the \emph{h} score as an objective metric to estimate it. In Section~\ref{sec:hdisc}, the discrimination model is described, together with possible shortcomings and a few alternative approaches. In the end, we conclude with the follow-up steps of the near future. 

\section{NLG Evaluation Trends}
\label{sec:nlgeval}

To have a quantitative picture of the NLG evaluation trends, we carefully examined the papers published in the last five INLG conference proceedings (2014--2019). We focused on the evaluation practices of these papers, sorting out those that used automatic metrics from those using human studies or those using both. The results are presented in Table~\ref{tab:evalstats}. 
As we can see, the number of studies performing automatic evaluation only (Auto column) has been rising, especially in the last year. Their portion goes up from 12\,\% in 2014 to 30.1\,\% in 2019. Among the evaluation metrics, BLEU was the most popular, followed by METEOR and ROUGE. The studies using both human and automatic evaluation (Both column) have also risen during the last five years. These findings are in line with earlier observations such as those of Gkatzia and Mahamood \cite{gkatzia-mahamood-2015-snapshot} who considered a wider range of NLP publications and reported an increase in automatic metric evaluations from 44\,\% in the period 2005--2008 to 60\,\% in 2012--2015.   
The opposite is true for the studies that perform human evaluation only (Human column). They have been following a decreasing trend, falling from 32\,\% to 11\,\%. A similar trend is reported in other recent surveys \cite{amidei-etal-2018-evaluation,8981519}. We also observed that for the human evaluation process, existing studies usually involve a few human experts or some dozens of students. Moreover, a considerable number of studies report to have used crowdsourced reviewers from Amazon Mechanical Turk or similar platforms. Accuracy, coherence, readability, and human likeliness are the most considered text quality criteria.   
Finally, there is also a category of studies (None column) that do not perform any evaluation or that do not assess the text generation quality. They were mostly proposals for shared tasks, findings from challenges, surveys or reviews, etc. 
%
\setlength\tabcolsep{3pt}
\begin{table}
\centering
\caption{\label{tab:evalstats}Evaluation types reported in the papers of the last five INLG proceedings (percentages in parenthesis).}
\begin{tabular}{|c|c|c|c|c|c|}
\hline
Event & Papers & Auto & Human & Both & None \\
\hline
12th INLG 2019 & 73 & 22\,(30.1) & 8\,(11) & 23\,(31.5) & 20\,(27.4)	 \\
11th INLG 2018 & 63 & 14\,(22.2) & 8\,(12.7) & 20\,(31.7) & 21\,(33.3) \\
10th INLG 2017 & 42 & 10\,(23.8) & 7\,(16.7) & 8\,(19) & 17\,(40.5) \\
~\,9th INLG 2016 & 44 & 12\,(27.3) & 10\,(22.7) & 7\,(15.9) & 15\,(34.1) \\
~\,8th INLG 2014 & 25 & 3\,(12) & 8\,(32) & 5\,(20) & 9\,(36) \\  
\hline
\end{tabular}
\end{table}

\section{Automating the Human Likeliness Evaluation}
\label{sec:human}

The results of our survey (and similar ones) suggest that it is highly desirable for researchers to benchmark the methods or techniques they propose using automatic metrics. As also pointed out by Reiter \cite{reiter-2018-structured} and Novikova et al. \cite{novikova-etal-2017-need}, 
new evaluation metrics are required. There is thus a strong incentive for formulating new quantitative measures and methodologies.   
Some attempts have developed objective measures like average word length, mean parse tree height, and the average number of nouns \cite{ambati-etal-2016-assessing,Vajjala.Meurers-14-eacl}. These measures are combined in formulas to obtain a score for automatically assessing text readability.
In this paper, we present a similar attempt regarding the human likeliness of the generated outputs. The \emph{h} score we propose reveals the capability of an NLG model to produce texts that are human-like or written by humans. This characteristic is important and highly desired in NLG applications as well as in various surging domains such as MT, TS, question answering and others that go beyond NLG. The most common approach is to assess human likeliness by means of human studies that ask participants to rate the texts using point-based schemes. Likert 5-points scale of is one of the most popular methods in the literature \cite{675159b2c1944257ba18910e46cd6dd1,van-der-lee-etal-2019-best}. 
Our idea is to automate the human likeliness evaluation of any text generation method \textbf{M} by considering the task as a binary discrimination problem and computing a metric that we call the \emph{h} score. Let's assume we are using a test set of $n$ reference samples for the evaluation. We can expect to have $n = n_h + n_m$, where $n_h$ is the number of texts that are perceived as human-written and $n_m$ is the number of those which are perceived as machine-generated. We can consider the \emph{h} score and \emph{m} score of \textbf{M} as the fraction of its outputs being perceived as human-written or machine-generated. They can be computed using Equation~\ref{eq:hscore}. 
\begin{equation}  
\label{eq:hscore}  
h^M = \frac{n_h}{n_h + n_m}	\qquad \textup{and} \qquad m^M = 1 - h^M = \frac{n_m}{n_h + n_m} 
\end{equation}
\noindent Instead of having human participants sorting out the texts (mark them as class \emph{h} or class \emph{m}) or giving them scores (e.g., 1 to 5 as in Likert scale), we propose to automate the process by using an intelligent discriminator model \textbf{H}. Using huge pretrained language models, it is possible today to automatically produce texts (e.g., news articles) that are almost impossible to distinguish. The success of this approach will thus depend on the possibility to create a smart \textbf{H} that is able to recognize such synthetic texts. Using the predictions of the discriminator and Equation~\ref{eq:hscore}, we then compute the \emph{h} scores of the methods we wish to evaluate and use it as a model quality indicator. We would thus favor the method which is more capable in fooling the discriminator to think that its text outputs are actually human-written (thus higher \emph{h} score).       

\section{The Human Likeliness Discriminator}
\label{sec:hdisc}

\subsection{GLTR and Pretrained Language Models}
\label{ssec:gltr}

The recent pretrained language models such as BERT of Devlin et al. \cite{devlin-etal-2019-bert} or GPT-2 of Radford et al. \cite{noauthororeditor} have led to significant advances in NLP research. Based on many transformer blocks and trained with huge volumes of texts, these models can be fine-tuned with specific data of various domains and yield top results in the corresponding tasks. 
It is possible to use big language models not just for synthetic text generation but also for spreading fake or misleading news, comments or reviews in the Web \cite{fornaciari-poesio-2014-identifying}. This has created strong incentives for research and development of synthetic text detection systems \cite{gehrmann-etal-2019-gltr,10.1145/3137597.3137600,perez-rosas-etal-2018-automatic}.
GLTR (Giant Language model Test Room) of Gehrmann et al. \cite{gehrmann-etal-2019-gltr} utilizes BERT, small GPT-2 (117\,M parameters) or large GPT-2 (1.5\,B parameters) models as backend for detecting synthetic texts of various sizes. Based on empirical observations, the authors assume that synthetic texts are mostly produced sampling words from the head of a language distribution model $p(X_i | X_{1:i-1})$ (high-rank words). To detect if a given word is likely sampled from a language distribution, they propose these tests: (i) checking the probability of the given word in relation to the one that was assigned the highest probability $P_{det}(X_i = \hat{X}_i | X_{1:i-1})$; (ii) checking the rank of the given word; (iii) checking the entropy of the predicted distribution. The higher these scores are, the higher are the chances that the given word and the entire text is synthetic (generated). They further build a visual tool that highlights text passages and can be used online.\footnote{http://gltr.io/dist/index.html} 

\subsection{The Discrimination Scheme}
\label{ssec:hdisc}

Our idea is to use the approach of GLTR for building the \textbf{H} discriminator described above. Big pretrained models such as BERT, GPT-2 small or GPT-2 large will provide the language model distribution $p(X_i | X_{1:i-1})$. Same as in \cite{gehrmann-etal-2019-gltr}, we assume that synthetic texts do mostly contain high-rank words and natural texts include more low-rank words. What remains is the designation of a numeric scheme by which \textbf{H} can compute and assess the quantity of high-rank words used in each sample and a discretization scheme to translate that quantity in a category (\emph{h} or \emph{m} class).    
We propose to compute $frac(p)$ for each word of the text sample (sequence $\hat{X}_{1:n}$). It is the fraction between the probability of a given word in its position and the highest probability of any word appearing in that position given by the language distribution of the pretrained model in use. From the $frac(p)$ values of each word, we can compute the average $frac(p)$ score ($Fp$) of a text sample $t$ consisting of $n$ words using Equation~\ref{eq:avgfp}: 
\begin{equation}  
\label{eq:avgfp}
fp_t = 1/n \sum_{i=1}^{n}\frac{P(\hat{X}_i)}{P(X_i)}
\end{equation}
As for the discretization, we will need a threshold or boundary value $Fp_b$ for the $Fp$ score and then sort out \emph{h} class samples from \emph{m} class ones using the scheme of Equation~\ref{eq:classes}:
\begin{equation}
\label{eq:classes}
class(t) =
 \begin{cases}
   h, & \text{if}\ Fp_t < Fp_b \\
   m, & \text{otherwise}
 \end{cases}
\end{equation} 
To set an optimal $Fp_b$ value, we will need to perform empirical examinations of many synthetic and natural texts and their respective $Fp$ scores. Based on our preliminary observations, $Fp_b$ should range somewhere between 0.35 and 0.45. After labeling each test sample output of our NLG method \textbf{M}, we can now use Equation~\ref{eq:hscore} to finally obtain its \emph{h} score. A high \emph{h} score reflects a high capability for producing samples that are perceived as \emph{h} class (texts with more low-rank words). It is thus an indication that the human likeliness of texts produced by \textbf{M} is high.

\subsection{Possible Flaws and Alternative Implementations}
\label{ssec:backup} 

Given that the $Fp$ values are continuous, using only $Fp_b$ as a boundary value may not be optimal. It may lead to a high missclassification rate of the texts. An alternative approach could be to use two threshold values for $Fp$: $Fp_l$ as a low boundary and $Fp_h$ as a high boundary. This way, we have a better separation of the two intervals for class \emph{h} ($0 < Fp < Fp_l$) and class \emph{m} ($Fp_h < Fp < 1$) by a third interval ($Fp_l < Fp < Fp_h$) that constitutes the \emph{u} (for \emph{unknown}) class of texts. In other words, a better approach could be to use Equations~\ref{eq:classes2} and \ref{eq:hscore2} instead of Equations~\ref{eq:classes} and \ref{eq:hscore}.
\begin{equation}
\label{eq:classes2}
class(t) =
\begin{cases}
h, & \text{if}\ Fp_t < Fp_l \\
u, & \text{if}\ Fp_l < Fp_t < Fp_h \\
m, & \text{otherwise}
\end{cases}
\end{equation} 
\begin{equation}  
\label{eq:hscore2}
h^M = \frac{n_h}{n_h + n_m + n_u} \qquad \textup{and} \qquad m^M = \frac{n_m}{n_h + n_m + n_u}
\end{equation}
\vskip 2mm
\noindent Once again, for finding the optimal $Fp_l$ and $Fp_h$ values, we will need to run several empirical tests using synthetic and natural samples. The $Fp_l$ and $Fp_h$ values will be set based on the average natural text $Fp$ and the average synthetic text $Fp$, together with their respective variances.
Another problem could be the loss of information from the discretizations schemes of Equations~\ref{eq:classes} or \ref{eq:classes2}. Discretizing the continuous $Fp$ and then computing the continuous \emph{h} score may result in a significant loss of precision. An alternative is to completely avoid the \emph{h} scores and instead use the $Fp$ values of the test samples to compute the average $Fp$ on the entire test dataset. This could be a cleaner practice, leading to a better quality indicator than the \emph{h} score. Moreover, this simpler approach could be adopted more easily, easing the cross-interpretation of the results.  
Another drawback is the fact that this approach can be easily fooled by word repetitions, grammatically incorrect words, etc. As a result, additional text checkups will be required to ensure its validity.   
Finally, the $Fp$ scores depend on the pretrained backend model that was used. In other words, when reporting the human likeliness of certain experiments, the backend used for the discrimination process should be reported as well.

\section{Discussion}
\label{sec:discussion}

In this paper, we first presented the results of a survey 
which reveals the trend towards the automatic evaluation of text quality criteria. We further described a method for automatically evaluating the naturalness of the NLG output texts by computing the human-likeliness score as the average of human-labeled test samples. Instead of relying on human participants to label those test samples, we propose to use a discrimination approach based on large pretrained language models like BERT or GPT-2 and the computation of the fraction probability of each word and the entire text. We also presented an alternative scheme that can be used to discretize the fraction probability values in case of high information loss from the discriminator.
Several empirical observations using synthetic and natural samples will be conducted to find the optimal setup of our proposal. The test set of the popular CNN/Dailymail news dataset created by Nallapati et al. \cite{K16-1028} is probably a good source of texts. Furthermore, we plan to conduct a comprehensive validation of the scheme by involving human participants who will judge the naturalness of the text samples. This will check the agreement between the automatic predictions and the human evaluations.

\bibliographystyle{splncs04}
\bibliography{bib}

\end{document}